%% file: main1.tex
\documentclass{article} %
\usepackage{iclr2025_conference,times}

\input{math_commands.tex}

\usepackage{hyperref}
\usepackage{url}

\usepackage{dirtytalk}
\usepackage{xspace}
\usepackage{wrapfig}
\usepackage{algorithm}
\usepackage{algpseudocode}
\usepackage{graphicx}

\newcommand{\says}[2]{\noindent\textcolor{purple}{\textbf{#1: }}\textcolor{purple}{#2}\xspace}

\newcommand{\sm}[1]{\says{Sam}{#1}}
\newcommand{\sg}[1]{\noindent\textcolor{orange}{\textbf{Shripad}: #1}}

\renewcommand{\sm}[1]{}
\renewcommand{\sg}[1]{}

\newcommand{\prob}{\mathbb{P}}
\newcommand{\N}{\mathbb{N}}
\renewcommand{\eps}{\varepsilon}
\newcommand{\Dpub}{D_{\text{pub}}}
\newcommand{\Dpriv}{D_{\text{priv}}}

\newcommand{\gem}{$\operatorname{GEM}^{\text{Pub}}$\xspace}

\newtheorem{theorem}{Theorem}[section]

\newtheorem{lemma}[theorem]{Lemma}

\newtheorem{definition}[theorem]{Definition}

\title{Leveraging Vertical Public-Private Split for Improved Synthetic Data Generation}

\author{Samuel Maddock, Shripad Gade, Graham Cormode \& Will Bullock \\
Meta Platforms, Inc.\\
\texttt{\{smaddock,shripadgade,gcormode,bullock\}@meta.com} \\
}

\iclrfinalcopy %
\begin{document}

\maketitle

\begin{abstract}
Differentially Private Synthetic Data Generation (DP-SDG) is a key enabler of private and secure tabular-data sharing, producing artificial data that carries through the underlying statistical properties %
of the input data. This typically involves adding carefully calibrated statistical noise to guarantee individual privacy, at the cost of %
synthetic data quality.
Recent literature has explored scenarios where a small amount of public data is used to help enhance the quality of synthetic data. These methods study a horizontal public-private partitioning which assumes access to a small number of public rows that can be used for model initialization, providing a small utility gain. 
However, realistic datasets often naturally consist of public and private attributes, making a \textit{vertical public-private partitioning} relevant for practical synthetic data deployments. We propose a novel framework that adapts horizontal public-assisted methods into the vertical setting. We compare this framework against our alternative approach that uses conditional generation, highlighting initial limitations of public-data assisted methods and proposing future research directions to address these challenges.

\end{abstract}

\section{Introduction}\label{sec:intro}

Due to increasing demand for privacy-preserving data sharing, differentially private synthetic data generation (DP-SDG) has emerged as a powerful tool for sharing sensitive tabular data. By generating \say{artificial} data that retains the statistical properties of the real data and adhering to formal privacy constraints, DP-SDG allows organizations to share, analyze and train models on data without exposing sensitive information \citep{assefa2020generating, van2023beyond}. Differential Privacy (DP) outlines a formal framework for individual privacy, providing SDG methods with quantifiable privacy guarantees, often at the expense of synthetic data quality \citep{dwork2006calibrating, hardt2012simple}. State-of-the art (SOTA) methods for private tabular synthetic data use DP to offer strong privacy guarantees. These methods typically involve taking low-dimensional measurements of the dataset, such as marginal queries, and adding calibrated statistical noise to ensure DP. A model is trained over these noisy measurements to learn a synthetic data representation that can be sampled from, such as a graphical model \citep{zhang2017privbayes, mckenna2019graphical} or generator neural networks \citep{liu2021iterative}.

Many real-world datasets naturally contain a mix of public and private information which can be leveraged to improve synthetic data quality (utility). In the context of differentially private learning, the concept of public-data assisted algorithms is well-studied \citep{bie2022private, ganesh2023public, ullah2024public}. Most existing DP-SDG algorithms treat the input data as entirely private \citep{zhang2017privbayes, zhang2021privsyn, mckenna2022aim}. However, real-world tabular datasets often contains a natural partition of public and private information. For example, demographic statistics like age may be publicly available from census information, while sensitive attributes related to financial or medical data require strict privacy protections. In many industry applications, companies that collect sensitive information have subsets of data that can be freely used, e.g., information users have explicitly consented to be used for improving products which can be used by the SDG algorithm. 

Recent research has introduced methods for synthetic data generation in public-data assisted settings, primarily focusing on a \textit{horizontal} partitioning  \citep{liu2021leveraging,liu2021iterative}. In this setting, a subset of rows over the entire dataset is considered public while the remainder is private. These methods adapt existing marginal-based DP-SDG algorithms to make use of the public rows in order to improve synthetic data quality \citep{wang2023post}. In several cases, the essence of these methods is to simply initialize the synthetic data model with public data, then proceed with training on the private dataset under DP. Leveraging the public subset in this way has been shown to enhance the utility of DP synthetic data \citep{liu2021iterative}. While this approach can improve synthetic data quality, it ignores the public data after initialization (risking forgetting), and so there is potential for more closely integrating public data into the private algorithm. Furthermore, it is often unrealistic to assume that the public subset is identical in distribution to the private dataset:  in many scenarios the public subset will be small and from a biased population, so may result in a poor model initialization. Meanwhile, if it were the case that the public subset does represent the private distribution accurately, then there is little benefit to be gained from fine tuning on the private data.  

In this work, we study an equally important but unexplored problem of a \textit{vertical} public-private partitioning. Here, some subset of columns in the dataset are considered public, while the rest are treated as private. This vertical partitioning is common in practice, as organizations frequently handle datasets with well-defined public and private features. In addition, these applications often involve scenarios where the number of private columns are much smaller than the number of public columns, necessitating the use of public-data assisted DP-SDG in order to obtain high utility. Despite its practical relevance, the vertical setting remains under-explored in the literature. Our contributions are as follows:
\begin{itemize}
    \item \textbf{Vertical Public-Private Setting}: We propose a framework for the adaptation of horizontal public-assisted DP-SDG methods to the vertical setting. We find our methods achieve good utility whilst existing baselines often fail to improve over fully private methods, even in settings where the percentage of public columns is large.
    \item \textbf{Conditional Generation:} We propose an alternative approach based on conditional generation, adapting marginal-based methods that use Private-PGM. We find this approach greatly improves synthetic data quality in a vertical setting and achieves best utility overall.
    \item \textbf{Re-thinking public-private SDG:} We enumerate the current limitations and propose future directions to help advance public-assisted private synthetic data generation in the vertical setting.
\end{itemize}

\section{Related Work}\label{sec:related_work}

A key class of methods for generating private synthetic data are marginal-based methods \citep{liu2021iterative}. These methods primarily focus on learning a synthetic data distribution by measuring low-dimensional statistics, such as marginal queries, under DP noise. Examples include PrivBayes \citep{zhang2017privbayes}, which learns a Bayesian network from noised mutual information, and Private-PGM, which trains a graphical model over noisy marginals \citep{mckenna2019graphical}. Approaches using neural networks have gained popularity with methods like DP-CTGAN \citep{fang2022dp}, PATE-GAN \citep{jordon2018pate}, and DP-VAE \citep{weggenmann2022dp} which employ deep learning techniques to directly model synthetic data and are trained privately via DP-SGD \citep{abadi2016deep}. Our focus in this work is on private marginal-based methods, as they are shown to outperform their neural network counterparts on tabular data \citep{liu2022utility, ganev2023understanding}.

Recent advances in marginal-based methods follow the \say{select-measure-generate} paradigm \citep{liu2021iterative}. The goal is to generate synthetic data that preserves the answers to a given workload of queries. The approach is iterative and involves a number of steps. In each iteration, the query that is worst-approximated under the current synthetic model is selected under DP. This marginal is measured and perturbed with DP noise, and is used to update the synthetic model. The MWEM \citep{hardt2012simple} algorithm was the first to adopt such an approach, but its synthetic data representation directly models the entire joint distribution leading to poor scalability as the number of columns increase. Alternative methods like GEM \citep{liu2021iterative} replace the inefficient representation in MWEM with a generator neural network that is trained directly on noisy measurements. The current SOTA is AIM \citep{mckenna2022aim}, which combines  Private-PGM \citep{mckenna2019graphical}, a graphical model inference procedure on noisy measurements, with a more sophisticated selection algorithm in order to achieve high utility and scale to a larger number of columns. 

The integration of public data for improving the utility in DP-SDG algorithms has been explored under a horizontal partitioning. In such settings, a small subset of rows is assumed to be public. These methods either pretrain the synthetic data model on public data before using the (fully) private algorithm or augment the training process of the private SDG algorithm with statistics from the public subset. \cite{liu2021leveraging} propose PMW, a public-data assisted version of MWEM. They initialize both the synthetic data distribution and its support using public data, then proceed as normal with MWEM. They show the use of public data can help bridge the gap between the fully private setting and perfectly fitting the training data. However, their focus is restricted to a horizontal partitioning and uses MWEM, which is a method that suffers on high-dimensional data. \cite{liu2021iterative} propose a public-data assisted version of GEM. The idea is analogous to PMW: the generator network in \gem is pretrained on all marginals in the public dataset, then the \gem algorithm proceeds as normal on the private dataset, using the pretrained model as initialization. This brings both added utility and scalability over PMW. \cite{wang2023post} propose a post-processing scheme that can be applied to the output of any SDG to help improve utility, including the public-data assisted setting. Finally, \cite{fuentes2024joint} propose JAM-PGM, a public-data assisted version of the AIM algorithm. The selection step of AIM is extended to decide at each iteration whether to measure a private marginal (with DP noise) or use a public (noise-free) marginal. This more closely involves the public data in the training process %
and helps achieve better utility than \gem in practice. 

The only other work we are aware of that addresses a vertical partitioning for public-assisted DP-SDG is due to \cite{liu2021iterative}, who outline extensions to \gem that allow a vertical partitioning. In essence, since \gem trains a generator network over marginals, it can be pretrained on any set of public measurements, computed from either a horizontal or vertical partitioning.

\section{Methods}\label{sec:method}

We assume access to a dataset $D$ over $n$ individuals, vertically partitioned into a private dataset $\Dpriv \in \N^{n \times d_{\text{priv}}}$ and a public dataset $\Dpub \in \N^{n \times d_{\text{pub}}}$ with $d := d_{\text{priv}} + d_{\text{pub}}$ being the total number of columns in $D$\footnote{We assume the dataset is discrete as in prior work. In practice, numerical features can be binned.}. We are mainly interested in the setting where $d_{\text{priv}} < d_{\text{pub}}$. The goal is to generate a synthetic dataset $\hat D \in \N^{n \times d}$ under differential privacy. In this work we are concerned with guaranteeing $(\eps, \delta)$-DP. 
\begin{definition}[$(\varepsilon, \delta)$-DP] 
A randomized algorithm $\mathcal{M}$ is $(\eps, \delta)$-DP if for any neighboring datasets $D, D^\prime$ and any subset of outputs $S$ we have
    $\prob(M(D) \in S) \leq e^\eps \prob(M(D^\prime) \in S) + \delta$
\end{definition}
The parameter $\eps$ is the privacy budget and determines how strong the privacy guarantee is. Using a large $\eps$ will decrease the noise in \say{Select-Measure-Generate} algorithms but in turn reduces the formal privacy guarantees. The parameter $\delta$ determines a probability that the DP guarantee fails and is usually set to be cryptographically small. Alternative definitions seeks to improve composition results such as zero-Concentrated Differential Privacy (zCDP)~\citep{bun2016concentrated}. All methods in our experiments use zCDP. See Appendix \ref{appendix:dp} for full details. In the vertical public-assisted setting we consider in this work, the DP guarantee of $\mathcal{M}$ is applied to $\Dpriv$ and we assume $\mathcal{M}$ has access to $\Dpub$ for no additional privacy cost (as it is public information).

The general vertical public-assisted DP-SDG framework is summarised in Algorithm \ref{alg:main}. Here we adapt the adaptive measurements framework proposed by \cite{liu2021iterative} to our vertical public-assisted setting. There are two key areas that differ. Firstly, model initialization may depend on $\Dpub$ i.e., for public pretraining. Secondly, the measurement step may also depend on $\Dpub$ e.g., for measuring a public marginal which requires no DP noise. \sm{dp def? exposition of zcdp accounting? marginal query def?}

\subsection{Adapting Pretraining Methods to the Vertical Setting}
\label{sec:pretrain}

Existing DP-SDG methods can be adapted to the vertical public-assisted setting by fitting them into the framework outlined in Algorithm \ref{alg:main}. \sg{Do you want to claim a unifying framework for public-data assisted DP-SDG? Also if this can be perfectly fit, we could create a table with each algo being a column and each row being a subroutine in the framework, e.g. TRAIN-MODEL. We should also be clear about where the extension was previously reported vs adapted by us.}\sm{I think a table could be good but we will have to see if there is enough space for this after experiments are put in.}

\paragraph{v\gem \citep{liu2021iterative}:} The GEM method uses a generator neural network to model the synthetic data distribution. No pre-processing of the workload is used, so $W^* = W$. The model initialization in the vertical setting pre-trains the generator neural network on all 3-way marginals in $W$ that contain only public features in $\Dpub$. %
Next, the error scores for query selection simply measure the error between the current synthetic data model and the private dataset at step $t$, with $\text{SCORE}(q; D, \hat D_t) := \|M_{q_t}(D) - M_{q_t}(\hat D_t)\|$. $\text{TRAIN-MODEL}(\cdot)$ performs a number of SGD steps on the current GAN $\theta_t$ to produce $\theta_{t+1}$. The gradients for these updates are computed from the average $L_1$ loss between the current marginals produced by $\theta_t$ and the observed noisy measurements $\{\tilde M_1, \dots, \tilde M_t \}$. Finally, the post-processing function $f(\cdot)$ performs an Exponential Moving Average (EMA) over the last $T/2$ generator networks, and the final synthetic dataset $\hat D$ is sampled from this.

\begin{algorithm}[t]
    \caption{Vertical Public-assisted Adaptive Measurements (vPAM)}\label{alg:main}
    \begin{algorithmic}[1]
        \Require Private dataset $\Dpriv$, public dataset $\Dpub$, workload of queries $W$, training steps $T$, privacy parameters $(\eps, \delta)$
        \Ensure Synthetic data $\hat D$
        \State Pre-process the workload $W^* := \text{PROCESS-WORKLOAD}(W)$ %
        \State $\theta_0 := \text{MODEL-INIT}(\Dpub)$, $\hat D_0 \sim \theta_0$ %
        \For{$t=0, \dots, T-1$}
            \State \textbf{Select:} via the Exponential mechanism $q_{t+1} \in W^*$ using $\text{SCORE}(q; \Dpriv, \Dpub, \hat D_t)$ %
            \State \textbf{Measure:} selected marginal query $q_{t+1}$ i.e., $\tilde M_{t+1} := \text{MEASURE}(q_{t+1}; \Dpriv, \Dpub, \sigma^2)$ %
            \State \textbf{Update:} synthetic model $\theta_{t+1} := \text{TRAIN-MODEL}(\theta_t, \{\tilde M_1, \dots, \tilde M_{t+1}\})$ %
            \State \textbf{Generate:} $\hat D_{t+1} \sim \theta_{t+1}$
        \EndFor
        \State Output $\hat D \sim f(\{\theta_t\}_{t=1}^T)$ %
    \end{algorithmic}
\end{algorithm}

\paragraph{vPMW:} PMW can be adapted to the vertical setting in a similar way to v\gem. The pretraining process for MODEL-INIT is the same as in v\gem, where the initial model $\theta_0$ is initialized over all 3-way marginals in $W^*$ that only contain columns in $\Dpub$. All other steps remain the same as v\gem except for $\text{TRAIN-MODEL}(\cdot)$, which replaces the generator neural network via direct modelling using multiplicative weights i.e., $\theta_{t+1} := \theta_{t} \cdot \exp(q_{t+1}(x) \cdot (\tilde M_{t+1} - q_{t+1}(\theta_{t}))/2n) $. Finally, the post-processing function averages all $T$ synthetic distributions, $f(\cdot) := \frac{1}{T} \sum_t \theta_t$. 

\subsection{Adapting JAM-PGM to the Vertical setting}
\label{sec:jam}

\paragraph{JAM-PGM \citep{fuentes2024joint} :} The original algorithm extends AIM to the horizontal public-assisted setting. Firstly, JAM-PGM initializes a workload $W^*$ that has two additional properties: it contains the downward closure of $W$ (i.e., all lower order marginals that can be formed from queries in $W$ are also added to $W^*$) and all public-marginals that can be measured from $\Dpub$ are added to $W^*$ separately from those that can also be measured on $D$. The scoring functions are adapted from PMW to take into account the predicted error of selecting a public or private marginal. More specifically, the score for a private marginal is $\text{SCORE}(q; D, \hat D_t) :=  \|M_{q_t}(D) - M_{q_t}(\hat D_t)\| - \sqrt{2/\pi}\sigma n_{q_t}$ which adapts the PMW score to include an expected error term based on measuring $q_t$ under Gaussian noise. For the case of a public marginal, the score is adapted to $\|M_{q_t}(D) - M_{q_t}(\hat D_t)\| - \frac{|D|}{|\Dpub|}\|M_{q_t}(D) - M_{q_t}(\Dpub)\|$ which replaces the expected error of Gaussian noise with the expected error of measuring from the smaller public dataset. The measurement step is changed to adapt to $W^*$, where if $q_t$ is public, then the marginal is measured without noise on $\Dpub$ i.e., $\tilde M_t := q_{t}(\Dpub)$ otherwise the measurement happens under Gaussian noise as normal, $\tilde M_t := q_t(D) + N(0, \sigma^2)$. The initialization step for JAM-PGM uses a random graphical model and does no public pretraining. Finally, the post-processing function $f(\cdot)$ returns the last model $\theta_T$, since a new graphical model is estimated at each step with (potentially) different structures.

\paragraph{vJAM-PGM:} When adapting JAM-PGM to the vertical setting, the augmented workload $W^*$ still includes both private and public marginals, except now the same marginal is only added to the workload once as it is either deemed private (contains at least one private column) or public (contained within $\Dpub$). This necessitates changing the score functions, since now the expected error of measuring a public marginal is zero, as it is exact. When measuring a public marginal, we use the original MWEM scoring i.e., $\text{SCORE}(q_{\text{public}}; \Dpub, \hat D_t) := \|M_{q_t}(\Dpub) - M_{q_t}(\hat D_t)\|$. Furthermore, we initialize the graphical model on all $1$-way marginals as in the original AIM algorithm \citep{mckenna2022aim}, except the public columns from $\Dpub$ are used as-is without any Gaussian noise. \sg{Is the Pragmatic AIM now vertical JAM-PGM? } \sm{Essentially yes.}

\subsection{Conditional Generation}
\label{sec:cond}
In the vertical setting we have access to the public columns ($\Dpub$) of the underlying dataset. We can naturally use this to improve the synthetic data sampling process. We consider conditional generation for marginal-based methods that utilize the Private-PGM algorithm \citep{mckenna2019graphical}, whose synthetic data representation is a graphical model. In this process, the graphical model forms a factorisation of the joint probability distribution from which, given an elimination order, we can sample synthetic data. To adapt this to a vertical partitioning, during the generation/sampling phase we simply use the raw data for public columns (i.e., exact marginals) and only sample private columns via the graphical model. %

\section{Experiments}\label{sec:experiments}

\paragraph{Methods.} We compare (fully) private AIM \citep{mckenna2022aim} against the closest baseline \gem \citep{liu2021iterative} which we denote v\gem and our two vertical approaches: vPMW, vJAM-PGM. We train these methods on a workload of all $3$-way marginals. For methods using Private-PGM (AIM, vJAM-PGM) we apply our conditional generation approach. For conditional methods, since the public columns are generated as is, we remove them from the training workload and instead use a workload of $3$-way marginals that contain at least one private column.

\paragraph{Datasets.} We focus on two public datasets. The first is Adult \citep{adult}, used in prior horizontal public-private work \citep{liu2021iterative, fuentes2024joint} and also by \gem in a vertical setting \citep{liu2021iterative}. As PMW cannot scale to more than a few columns, we also consider Adult (red.), a smaller version containing only the first $8$ columns. The second dataset we use is the Census-Income KDD dataset \citep{census-income} which we use as a proxy for a more practical large-scale dataset allowing us to evaluate our methods across a larger range of public-private splits. We filter out rows with missing values to obtain 95,130 rows and $40$ columns.

\begin{wrapfigure}[16]{r}{0.5\textwidth}
    \vspace{-9mm}
  \begin{center} 
    \includegraphics[width=0.49\textwidth]{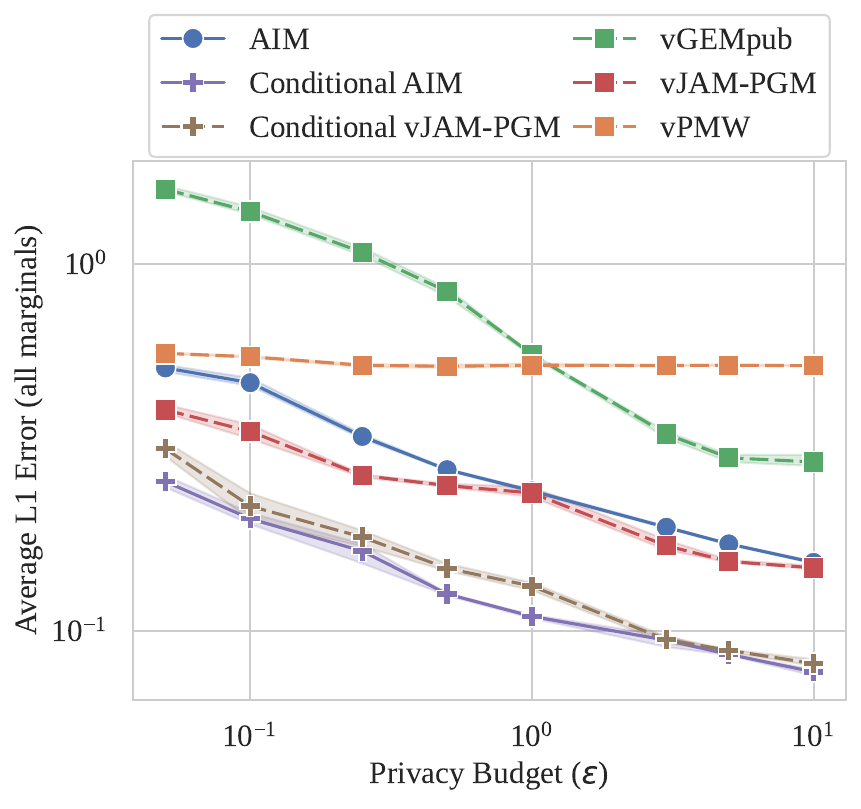}
  \end{center}
  \vspace{-5mm}
  \caption{Varying $\eps$ on Adult (red.). with $d_{\text{pub}} = 6$ \label{fig:1}}
\end{wrapfigure}

\paragraph{Evaluation.} We repeat experiments $3$ times and present the average L1 error over all $k$-way marginals with $k \leq 3$. See Appendix \ref{appendix:impl} for full experiment hyperparameters and open-source implementations used.

\subsection{Results}\label{sec:results}

\paragraph{Comparison of methods on Adult (reduced).} We start by comparing all methods on Adult (red.), taking $75\%$ of the columns to be public whilst varying the privacy budget $\eps$, shown in Figure \ref{fig:1}. We observe, like in prior horizontal work, that PMW has generally poor utility across all values of $\eps$. We also find v\gem struggles to achieve good utility, performing worse than (fully) private AIM across all settings. For methods based on Private-PGM, we see that when $\eps < 1$, vJAM-PGM achieves consistently lower error than AIM but that as $\eps$ grows large this gap starts to decrease, since the benefit of using public columns diminishes. We omit vPMW from further experiments, due to poor scalability and utility. 

\begin{figure*}[t]
    \centering
    \includegraphics[width=\textwidth]{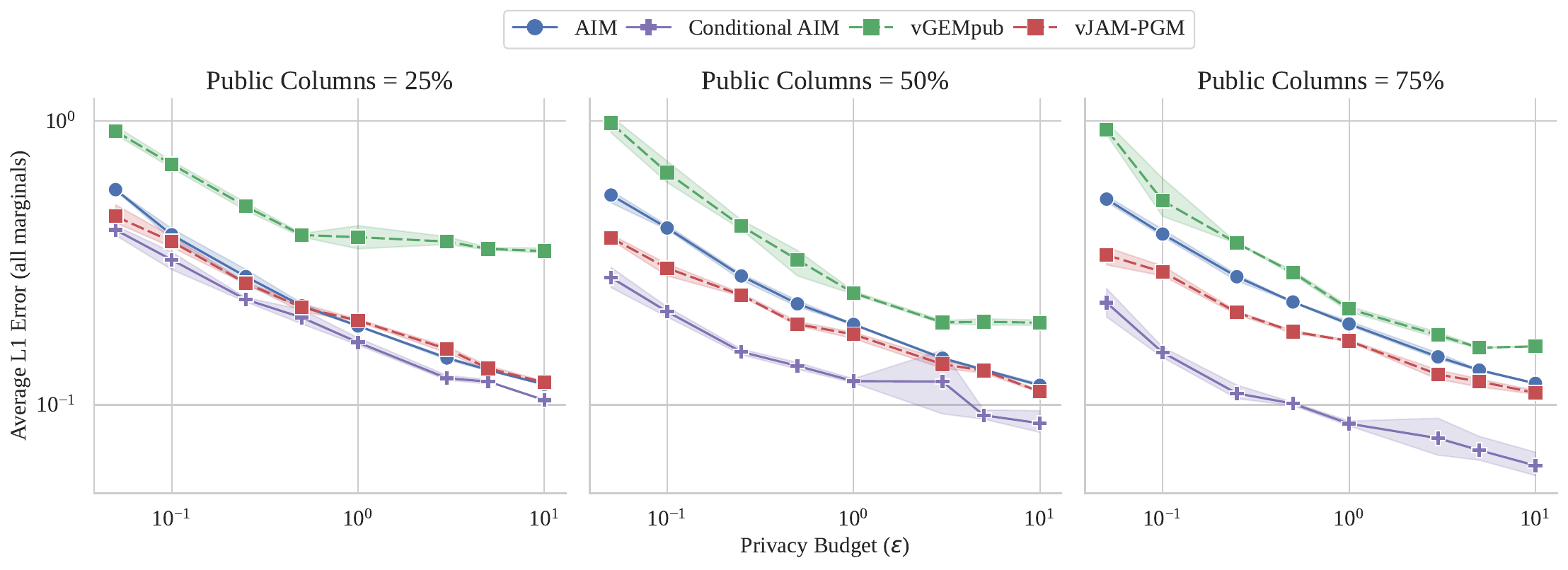}
    \caption{Varying $\eps$ on Adult with [$25\%, 50\%, 75\%$] of the columns being public. \label{fig:2}}
\end{figure*}

\paragraph{Varying $\eps$.} In Figure \ref{fig:2}, we vary $\eps$ on the full Adult dataset with 15 columns, taking [$25\%$, $50\%$, $75\%$] of the columns to be public. When $25\%$ of the columns are public, none of the vertical public-assisted methods provide better utility than using (fully) private AIM except when $\eps < 0.1$ for vJAM-PGM. This changes as the number of public columns increases, and at $50\%$ with $\eps < 1$ vJAM-PGM achieves lower error than AIM. For 75\% public, vJAM-PGM achieves lower error than AIM across all values of $\eps$. We note v\gem has consistently worse error than (fully private) AIM across all settings.

\begin{figure*}[t]
    \centering
    \includegraphics[width=0.8\textwidth]{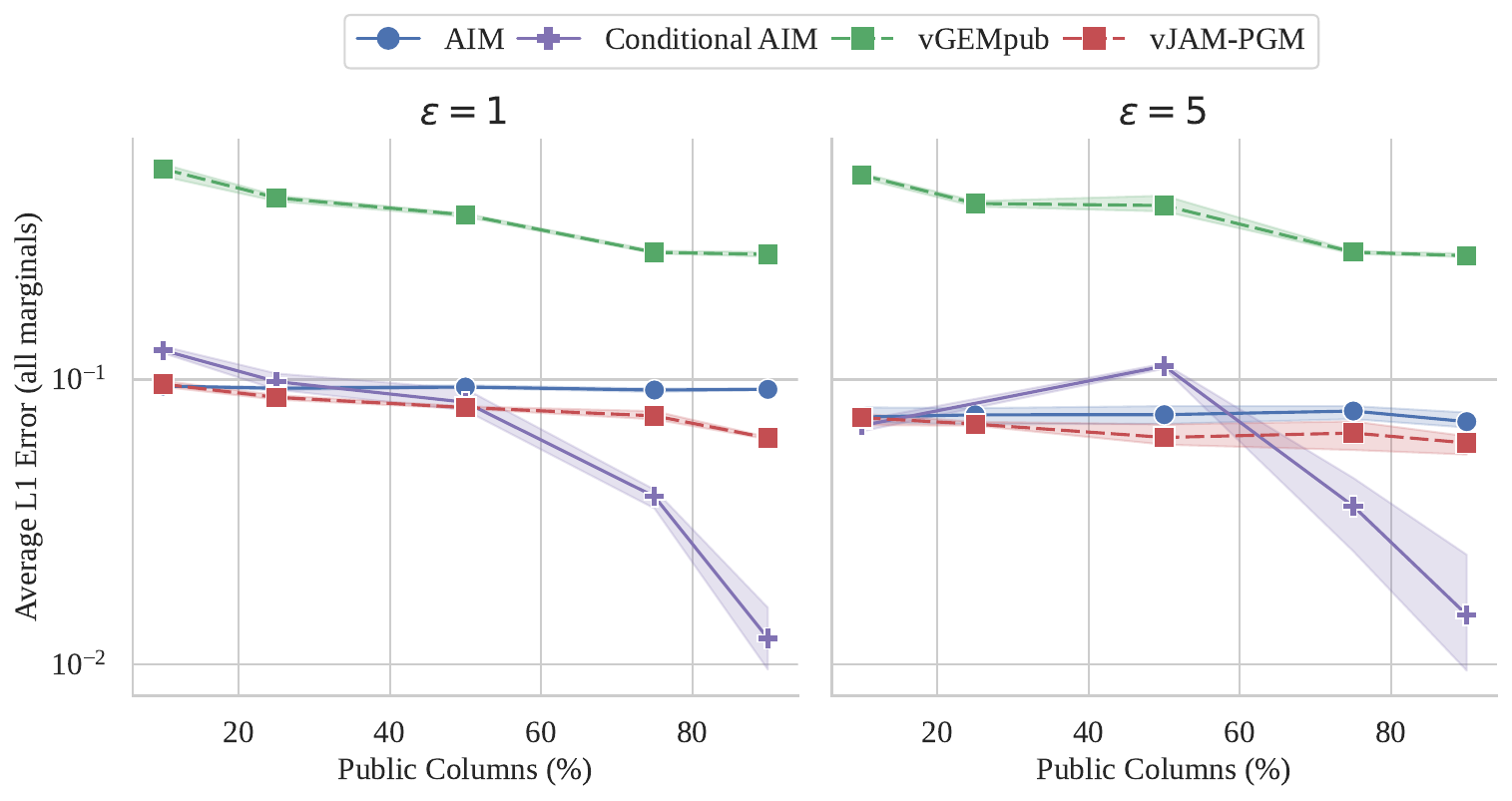}
    \caption{Varying the percentage of public columns on Census data with $p \in \{10\%, 25\%, 50\%, 75\%, 90\%\}$ and $\eps=1,5$ 
    \label{fig:3}}
\end{figure*}

\paragraph{Conditional Generation.} In Figure \ref{fig:1} we plot conditional variations of AIM and vJAM-PGM. We find both of these methods achieve superior utility, outperforming all other vertical methods. We omit Conditional vJAM-PGM from further experiments, since it performs consistently worse than Conditional AIM. Figure \ref{fig:2} and \ref{fig:3}, further shows Conditional AIM achieving lowest error across all settings, even beating AIM in scenarios where the number of public columns is small. The advantage of conditional generation not only helps achieve zero error across all public-marginals but consistently lowers private error as well. 

\paragraph{Varying the public-private split ($p$).} In Figure \ref{fig:3}, we consider a high-privacy ($\eps = 1$) and low-privacy ($\eps = 5$) setting on the Census dataset where we vary the percentage of public columns. %
We find, as on Adult, that vertical methods only achieve lower utility than (fully) private AIM when the number of public columns is $> 50\%$. We continue to observe the benefit of conditional generation, where the gap against vJAM-PGM is most striking when $p \geq 75\%$.


\section{Discussion}\label{sec:discussion}

\paragraph{Utility.} A consistent finding from our experiments is that vertical public-assisted methods  struggle to remain competitive against (fully) private AIM. 
That is to say, the additional pretraining on public data currently has little effect on the overall utility. The exception is vJAM-PGM, which integrates public data more closely into the training of the model, but only beats AIM when either $\eps$ is small or $d_{\text{pub}}$ is large. The only approach with consistently low error is Conditional AIM.

\paragraph{Scalability.} Although Conditional AIM achieves best utility, we have found scenarios where conditional sampling can require an intractable amount of memory when conditioning on a large number of public columns. We have found that heuristic approaches based on restricting the size of public marginals used in conditional sampling can alleviate these issues with no significant changes to utility. We note these problems are inherent to methods based on Private-PGM which struggle to scale to a large number of columns in both the fitting and generation step. This is prohibitive in industry applications 
where high-dimensional data %
is ubiquitous and is an area where we hope to make improvements in the future.

\paragraph{Future Directions.} 
We have found that adaptations of existing methods do not always provide sufficient benefit over (fully) private AIM, even when there are more public than private columns. Instead, we observe that moving away from pretraining approaches and towards conditional generation can help achieve the best synthetic data quality in the vertical setting. We believe there are many future directions for improving public-assisted methods. One such focus is enhancing both the scalability and utility of conditional methods, by improving the elimination order for Private-PGM based methods or adapting the generator networks in GEM to allow for conditional generation. Another direction is to focus the design of future vertical public-assisted algorithms away from pretraining and to instead make more use of public data during training such as in vJAM-PGM.

\bibliography{iclr2025_conference}
\bibliographystyle{iclr2025_conference}

\newpage
\appendix
\section{Differential Privacy}
\label{appendix:dp}
In this work we are concerned with synthetic data methods that guarantee $(\eps, 
\delta)$-DP. 
\begin{definition}[$(\varepsilon, \delta)$-DP] 
A randomized algorithm $\mathcal{M}$ is $(\eps, \delta)$-differentially private if for any neighboring datasets $D, D^\prime$ and any subset of outputs $S$ we have
    $\prob(M(D) \in S) \leq e^\eps \prob(M(D^\prime) \in S) + \delta$
\end{definition}
We consider neighboring datasets to mean that $D^\prime$ is formed from the addition or removal of a single data-point in $D$. One useful property of DP is composition, that is the output of two DP algorithms on $D$ with privacy parameters $\eps_1, \eps_2$ is $(\eps_1 + \eps_2)$-DP. Alternative DP definitions seek to improve these composition results. One such approach is zero-Concentrated Differential Privacy (zCDP)\citep{bun2016concentrated}.

\begin{definition}[$\rho$-zCDP] \label{def:zcdp}
An algorithm $\mathcal{M}$ is $\rho$-zCDP if for any two neighbouring datasets $D, D^\prime$ and all $\alpha \in (1,\infty)$ we have $D_\alpha(\mathcal{M}(D) | \mathcal{M}(D^\prime) \leq \rho \cdot \alpha$,
where $D_\alpha$ is Renyi divergence of order $\alpha$.
\end{definition}

We can convert an associated $\rho$-zCDP guarantee into $(\eps, \delta)$-DP via the following lemma.

\begin{lemma}[zCDP to DP \citep{canonne2020discrete}]
\label{lemma:cdp}
If an algorithm $\mathcal{M}$ satisfies $\rho$-zCDP then it satisfies $(\eps,\delta)$-DP for all $\eps > 0$ with $$\delta = \min_{\alpha > 1} \frac{\exp((\alpha-1)(\alpha\rho - \eps))}{\alpha - 1} \left(1-\frac{1}{\alpha}\right)^\alpha$$
\end{lemma}

Existing methods like AIM\citep{mckenna2022aim} utilize zCDP accounting.  We implement all vertical \say{Select-Measure-Generate} methods in our framework with zCDP accounting, composing the Exponential (\say{Select} step) and Gaussian (\say{Measure} step) mechanisms over $T$ rounds. We note the privacy accounting of vJAM-PGM following that of JAM-PGM spends some extra privacy budget to decide whether to pick a public or private marginal at a given round under the framework of the Exponential mechanism. We refer to the original work for full privacy details \citep{fuentes2024joint}.

\begin{definition}[Gaussian Mechanism]
Let $q: \mathcal{D} \rightarrow \R^d$ be a sensitivity $1$ query, the Gaussian mechanism releases $q(D) + \Delta_2(f) \cdot \mathcal{N}(0, \sigma^2 I_d)$ and satisfies $\frac{1}{2\sigma^2}$-zCDP.
\end{definition}
\begin{definition}[Exponential Mechanism]
Let $s(c) : \mathcal{D} \rightarrow \R$ be a score function defined over a set of candidates $C$. The exponential mechanism releases $c$ with probability $\mathbb{P}[\mathcal{M}(D) = c] \propto \exp(\frac{\eps}{2\Delta} \cdot s(c))$,
with $\Delta := \max_q \Delta_1(s(c))$. This satisfies $\frac{\eps^2}{8}$-zCDP.
\end{definition}

 Finally, we note that further accounting improvements can be made to improve composition when then number of rounds $T$ is large, using numerical accounting methods as in DP-ML \citep{mironov2017renyi, gopi2021numerical} but is not a setting we consider.

\section{Experiment Details}
\subsection{Implementation Details}
\label{appendix:impl}
In our experiments we use the following methods:
\begin{itemize}
    \item \textbf{vPMW and v\gem:} We use the open-source implementations of PMW and \gem by the original authors\footnote{\url{https://github.com/terranceliu/iterative-dp}}. We modify these methods to change the pretraining to only pretrain on marginals that contain public columns as discussed in Section \ref{sec:pretrain}.
    \item \textbf{vJAM-PGM:} We use the open-source implementation of JAM-PGM\footnote{\url{https://github.com/Miguel-Fuentes/JAM_AiStats}} by \cite{fuentes2024joint} with the modifications discussed in Section \ref{sec:jam}. 
    \item \textbf{AIM / Conditional AIM:} For AIM, we use the original implementation by the original authors\footnote{\url{https://github.com/ryan112358/private-pgm}}. For conditional AIM, we modify the Private-PGM algorithm to conditionally sample based on the public vertical columns as discussed in Section \ref{sec:cond}.
\end{itemize}

\subsection{Datasets and Hyperparameters}
We use the following datasets:
\begin{itemize}
    \item \textbf{Adult} \citep{adult}: We use the Adult dataset with $32,561$ rows and $15$ features. For Figure \ref{fig:1}, we use Adult (red.), a reduced version we form by using only the first $8$ columns.
    \item \textbf{Census-Income KDD} \citep{census-income}: We preprocess it by dropping all rows with missing values and any columns that are constant values. This gives a final dataset with $95,130$ rows and $40$ columns. We discretize numerical columns via (non-private) quantiles with a maximum of 50 distinct values. We note that this process can be done privately but is not the focus of our work.
\end{itemize}
Our experiments in Section \ref{sec:experiments} use the following hyperparameters:
\begin{itemize}
    \item \textbf{Privacy parameters ($\eps, \delta$)}: Whilst we vary $\eps$ in different experiments the privacy accounting is kept constant among methods in line with the discussion in Appendix \ref{appendix:dp}. We fix $\delta=1e-6$ in all experiments.
    \item \textbf{Number of Rounds ($T$)}: We fix this to $T = 100$ in all experiments. We varied $T$ and did not see any significant change to results. 
    \item \textbf{Pretraining Rounds}: For v\gem, we pretrain on all public marginals for $10$ rounds. For vPMW the synthetic data is initialized directly from the public marginals.
\end{itemize}
For other method specific hyperparameters (e.g., GEM learning rates or the number of iterations to optimize the graphical model in AIM) we use the same parameter defaults as the open-source implementations noted in Appendix \ref{appendix:impl}. This applies to v\gem, vPMW, AIM (also Conditional AIM) and vJAM-PGM.

\end{document}

%% file: math_commands.tex
\usepackage{amsmath,amsfonts,bm}

\def\eqref#1{equation~\ref{#1}}

\def\1{\bm{1}}

\def\eps{{\epsilon}}

\DeclareMathAlphabet{\mathsfit}{\encodingdefault}{\sfdefault}{m}{sl}
\SetMathAlphabet{\mathsfit}{bold}{\encodingdefault}{\sfdefault}{bx}{n}

\newcommand{\R}{\mathbb{R}}